\documentclass{article}
\usepackage[eandd,nonanonymous]{neurips_2026}[preprint] 
\usepackage[utf8]{inputenc} 
\usepackage[T1]{fontenc}    
\usepackage{hyperref}       
\usepackage{url}            
\usepackage{booktabs}       
\usepackage{amsfonts}       
\usepackage{nicefrac}       
\usepackage{microtype}      
\usepackage{xcolor}         
\usepackage{listings}
\usepackage{graphicx}
\usepackage{tcolorbox}

\usepackage[inline]{enumitem}

\lstdefinelanguage{prompt}{
    basicstyle=\ttfamily\small,
    stepnumber=1,
    numbersep=8pt,
    showstringspaces=false,
    breaklines=true,
    frame=single,
    backgroundcolor=\color{gray!10},
    string=[s]{"}{"},
    comment=[l]{//},
    morecomment=[s]{/*}{*/},
}

\title{UCSF-PDGM-VQA: Visual Question Answering dataset for brain tumor MRI interpretation}

\author{%
  Shiv Ghosh* \\
  Fung Institute for Engineering Leadership\\
  University of California, Berkeley\\
  Berkeley, CA 94709 \\
  \texttt{shiv.ghosh@berkeley.edu} \\
  \And
  Junayd Lateef* \\
  Fung Institute for Engineering Leadership\\
  University of California, Berkeley\\
  Berkeley, CA 94709 \\
  \texttt{jlateef@berkeley.edu} \\
  \And
  Chih-Hua Liu* \\
  Fung Institute for Engineering Leadership\\
  University of California, Berkeley\\
  Berkeley, CA 94709 \\
  \texttt{ch.liu@berkeley.edu} \\
  \AND
  Yannan Yu \\
  Department of Radiology\\
  University of California, San Francisco\\
  San Francisco, CA 94158 \\
  \texttt{yannan.yu@ucsf.edu} \\
  \And
  Andreas M. Rauschecker \\
  Department of Radiology\\
  University of California, San Francisco\\
  San Francisco, CA 94143 \\
  \texttt{andreas.rauschecker@ucsf.edu} \\
  \And
  Madhumita Sushil \\
  Division of Clinical Informatics and Digital Transformation and Department of Neurological Surgery\\
  University of California, San Francisco\\
  San Francisco, CA 94158 \\
  \texttt{Madhumita.Sushil@ucsf.edu} \\
}

\begin{document}

\maketitle
 
\begin{abstract}

Brain tumor diagnosis is largely dependent on Magnetic Resonance Imaging (MRI) evaluation, which requires radiologists to synthesize thousands of images across multiple 3D sequences and longitudinal studies. This process requires advanced neuro-radiology training, poses substantial cognitive load, and is highly time-consuming. Despite increasing demands in radiology, this expertise is difficult to scale, straining the current health systems. Vision-Language Models (VLMs) provide an opportunity to reduce this burden through a semi-automated, interactive interpretation of complex brain MRIs. However, they are currently underutilized in neuro-oncology due to a lack of specialized benchmarks for evaluating them. We introduce a clinically relevant visual question answering (VQA) benchmark --- the UCSF-PDGM-VQA dataset --- consisting of 2,387 QA pairs from 473 glioma-related MRI studies in the public UCSF-PDGM dataset. We further establish a performance baseline for six state-of-the-art vision-language models (VLMs) and one large language model on this dataset. We find that current models are incapable of effectively processing multi-sequence, 3-dimensional MRI scans, thus resulting in a suppression of visual features and over-reliance on language priors, causing modality collapse. These findings underscore a critical deficiency in current model reliability and safety within clinical settings, necessitating the development of robust, domain-specific VLMs.

\end{abstract}

\section{Introduction}
Primary brain and central nervous system (CNS) tumors impact more than 1 million individuals in the United States with an incidence of nearly 500,000 cases a year \citep{price2024cbtrus}. Brain tumors are a leading cause of death, with the median survival for an aggressive subtype called glioblastoma being only 12--18 months. Timely diagnosis and quick action are thus critical. Magnetic resonance imaging (MRI) is the primary imaging modality for detecting and monitoring brain tumors. However, interpreting brain tumor MRI scans is complex and time-consuming, as radiologists must analyze thousands of image slices, integrate information from multiple imaging sequences, and compare findings at multiple timepoints. On average, it takes 11--18 minutes to read a single brain MRI scan \citep{al2018time}, with more complex scans requiring hours. Prolonged periods of clinical image interpretation have been associated with a decrease in abnormality detection performance among radiologists \cite{Krupinski2010}. 

Recent advances in vision-language models (VLMs) are promising for developing an interactive radiology copilot to enable easier interpretation of complex scans. Despite the recent popularity of VLMs, as well as VQA with VLMs, their applications and use for complex diseases like brain tumors have been limited. Existing datasets for VQA are either limited to anatomies other than brain, or are artificially simplified, providing only 2-dimensional (2D) images when real-world studies instead require the processing of multiple 3-dimensional (3D) series. Clinical brain MRI studies routinely collect scans of types T1 pre-contrast, T1 post-contrast, T2, FLAIR, diffusion, SWI, and perfusion for accurate interpretation. Each scan type reveals a different aspect of the brain tumor by imaging distinct variations in tissue signal intensity to better view the brain tumor-associated abnormality. For example, T1 scans without contrast are used to understand the baseline anatomical view of the brain, T1 post-contrast scans provide information related to where the blood-brain barrier is disrupted or the tumor's vascularity is abnormal, and T2 FLAIR scans are useful for identifying edema and the boundaries of the tumor \citep{villanueva2017current}. A joint analysis of all these scan types is critical to ensure an accurate interpretation. Furthermore, new scans are often compared to prior scans to understand longitudinal changes associated with either disease progression or treatment response, which is critical to determine the next steps for the patient \citep{villanueva2017current}. The simplifications within existing datasets limit clinically significant progress, thus constraining the translation of VLMs or VQA into real-world clinical workflows.

To bridge this gap, we introduce a VQA dataset comprising clinically relevant concepts and scans, the UCSF-PDGM-VQA dataset, which provides a set of 2,387 closed-ended question-answer pairs answerable from 473 brain MRI studies for patients with diffuse gliomas. Each MRI study includes all imaging series collected during routine patient care, including the sequences described earlier. While the dataset includes scans only at a single time point --- preoperative scans, it is the first step towards developing a VQA dataset that reflects the complexity of real-world brain MRI processing. 

We further evaluated popular open-weight, clinical VLMs on this dataset to assess model performance. We identified key challenges in integrating all imaging series and thousands of slices in existing models - they are simply incapable of processing data of this complexity, resolution, and scale. Downsampling vision input to a few slices supported by existing models, we identified significant performance gaps; the best performing model, the MedGemma-1.5 model \citep{sellergren2025medgemma}, scored only 63.57\% accuracy on the task, with other models being significantly worse. Our ablation studies revealed a clear instance of modality collapse, wherein the multimodal models relied strictly on textual cues rather than visual data. This is evidenced by several VLMs improving when the imaging slice was replaced with a blank input. Notably, Lingshu-32B \citep{lasateam2025lingshugeneralistfoundationmodel} achieved the highest overall accuracy observed (66.04\%) when provided only with a blank image. This finding is further reinforced by the text-only LLM Qwen3-8B matching the performance of the multimodal MedGemma model. Alongside this modality collapse, the models exhibited a strong positional language bias: scores changed significantly when answer options were reordered despite identical inputs, confirming a reliance on structural heuristics rather than actual clinical reasoning. This highlights a key safety concern related to the use of VLMs in clinical settings: model responses are biased towards text prompts rather than the actual image demonstrating the disease presentation. This is deeply concerning, especially when, during a clinical review, the questions were deemed to be unanswerable without the accompanying radiology image. Together, these findings highlight the need for developing VLMs that are not only capable of processing multi-series 3D imaging, but also those that are robust to modality collapse to ensure that they can be safely translated for real-world clinical use. 



To summarize, our key contributions are as follows:
\begin{itemize}
\item We curated the first multiple-choice VQA dataset for brain tumor-associated MRI interpretation. The dataset will be made available on PhysioNet to enable future analysis on newer VLMs.
\item We benchmarked the capabilities of existing VLMs on this dataset, comparing model performance with clinical performance and identifying key performance gaps. All accompanying source code has been made available on Github \footnote{ \href{https://github.com/m2ai-lab/VLM-Brain-Tumor-QA-pipeline}{https://github.com/m2ai-lab/VLM-Brain-Tumor-QA-pipeline}}.
\item We evaluated vulnerabilities of existing VLMs on this VQA task, demonstrating the lack of effective integration of the vision modality, thus highlighting the risks associated with the clinical deployment of these models.
\item We designed a prototype graphical user interface enabling intuitive visualization and interaction with the underlying datasets and integrated models, hosted publicly at \textbf{redacted}.
\end{itemize}

\section{Related Work}

\subsection{Existing Biomedical VQA Datasets}

Several biomedical vision-language datasets are available publicly, enabling research on tasks such as biomedical VQA. Some notable VQA datasets include RadVisDial \cite{Kovaleva2019VisualDF}, Path-VQA \cite{he2020pathvqa}, VQA-Med (2021) \cite{imageclef_vqa_med2021}, MIMIC CXR VQA \cite{10.5555/3666122.3666292}, Medical-Diff-VQA \cite{PhysioNet-medical-diff-vqa-1.0.1}, VQA-RAD\citep{lau2018dataset}, SLAKE\citep{liu2021slake}, OVQA \citep{huang2022ovqa}, M3D-VQA \cite{bai2024m3d}, and PMC-VQA \cite{zhang2023pmc} (Table \ref{tab:vqa}), and additional vision-language datasets like CT-RATE \citep{Hamamci_2026} that do not provide a VQA subset. The brain is often not included as an anatomical structure in these datasets. Furthermore, many existing datasets are curated by retrieving images from public medical resources and repurposing image captions into question-answer pairs. Thus, 3D data, such as computed tomography (CT) and magnetic resonance imaging (MRI) exams, end up being represented as 2D, resulting in significant information loss and limiting their clinical utility. In contrast, real-world CT and MRI scans comprise thousands of 2D slices that together form a single 3D imaging series. Moreover, multiple series, such as pre-contrast and contrast-enhanced, and more advanced MRI-specific sequences, such as T2, FLAIR, and DWI, are jointly processed for reliable inference at a single time point. To our knowledge, the UCSF-PDGM-VQA dataset developed in this study is the first publicly available VQA dataset to encompass multiple brain anatomy imaging sequences, enabling VQA tasks in a more clinically aligned setting. Finally, questions included within existing datasets span topics such as imaging techniques, anatomical location, and organ systems visible, which are trivial for radiologists and not clinically meaningful for interpreting the scan \citep{mishra2025barriers}. In this study, we focus on VQA pairs derived from the findings and interpretations of the underlying radiology exam, creating a more clinically relevant dataset.

\begin{table}[h]
\centering
\caption{Comparison between existing and proposed clinical VQA datasets}
\resizebox{\textwidth}{!}{%
    \begin{tabular}{|l|l|l|l|l|l|l|l|}
    \hline
    \textbf{Dataset} & \textbf{Modalities} & \textbf{Anatomy} & \textbf{2D or 3D} & \textbf{Multi-series?} & \textbf{Size} & \textbf{QA format} & \textbf{Curation} \\
    \hline
    
    \hline
    \textbf{RadVisDial (2019)} & X-ray & Chest & 2D & No & 455,300 & Closed-ended & Automated \\
    \textbf{Path-VQA} & Pathology & Body Tissues & 2D & No & 32,799 & Open-, Closed-ended & Semi-automated \\
    \textbf{VQA-Med (2021)} & X-ray, CT, MRI & Breast, Skull, Face, Spine, Musculoskeletal & 2D & No & 5,000 & Open-ended & Semi-automated \\
    \textbf{MIMIC CXR VQA} & X-ray & Chest & 2D & No & 377,391 & Open-, Closed-ended & Automated \\
    \textbf{Medical-Diff-VQA} & X-ray & Chest & 2D & No & 700,703 & Open-ended & Semi-automated \\
    \textbf{VQA-RAD} & X-ray, CT, MRI & Head, Chest, Abdomen & 2D & Yes & 3,515 & Open-, Closed-ended & Semi-automated \\
    \textbf{SLAKE} & X-ray, CT, MRI & Head, Chest, Abdomen & 2D & No & 14,000 & Open-, Closed-ended & Manual \\
    \textbf{OVQA} & X-ray, CT & Head, Chest, Leg, Hand & 2D & No & 19,020 & Open-, Closed-ended & Semi-automated \\
    \textbf{M3D-VQA} & CT & Mainly Liver, Spleen, Kidney, Lung & 3D & No & 509,755 & Closed-ended & Semi-automated \\
    \textbf{PMC-VQA} & Mainly X-ray, CT, MRI & Brain, Lung, Heart, Liver & 2D & No & 226,946 & Closed-ended & Semi-automated \\
    \textbf{UCSF-PDGM-VQA} & MRI & Brain & 3D & Yes & 2,387 & Closed-ended & Semi-automated \\
    \hline
    
    \end{tabular}
}
\label{tab:vqa}
\end{table}

\subsection{Existing clinical vision-language models}

Several multimodal models, such as MedFlamingo \cite{moor2023med}, LLaVA-Med \cite{li2023llava}, BiomedCLIP \cite{doi:10.1056/AIoa2400640}, PubMedCLIP \cite{eslami2023pubmedclip}, MPMA \cite{zhang2023multi}, VILA-M3 \cite{nath2025vilam3enhancingvisionlanguagemodels}, Med3DVLM \cite{xin2025med3dvlm}, MedGemma-1.5 \cite{sellergren2025medgemma}, Lingshu \cite{lasateam2025lingshugeneralistfoundationmodel}, Merlin \cite{blankemeier_kumar2026merlin}, CT-CLIP \cite{Hamamci_2026}, Pillar-0 \cite{pillar0}, CALM-VLM \cite{Dhinagar2026.04.10.717865}, and Prima \cite{Lyu2026LearningNM}, have been recently developed for medical image analysis (Table \ref{tab:vlms}). These models typically build upon pretrained medical imaging and medical language backbones, which are subsequently refined and fine-tuned for specific imaging modalities or downstream tasks. Fusion strategies range from contrastive learning–based modality alignment to cross-modal attention and instruction tuning. However, despite continued domain-specific training, existing architectures remain limited in their ability to fully capture clinically relevant information. In most approaches, 3D medical images are resampled and standardized to a fixed tensor size, effectively reducing the volume to a predetermined number of slices before inference. While this standardization enables efficient model training, it can limit the volumetric context present in the original scan. Recent models such as Pillar-0 \citep{pillar0} and Prima \citep{Lyu2026LearningNM} attempt to address this limitation by processing complete 3D volumes; however, these approaches remain both anatomy- and modality-specific and are not yet trained for zero-shot VQA.
Table \ref{tab:vlms} provides a list of available clinical VLMs, and their key limitations for VQA over brain MRI data.

\begin{table}[h]
\centering
\caption{Existing clinical vision-language models and their key limitations for brain MRI VQA.}
\resizebox{\textwidth}{!}{%
    \begin{tabular}{|l|l|l|l|l|l|l|l|}
    \hline
    \textbf{Model} & \textbf{Vision Encoder} & \textbf{Text Encoder} & \textbf{Fusion Strategy} & \textbf{Key Limitation} \\
    \hline
    
    \hline
    \textbf{MedFlamingo} & CLIP ViT/L-14 &  LLaMA-7B \cite{touvron2023llamaopenefficientfoundation} & Gated Cross-Attention & Lacks neuro-oncology data, no VQA\\
    \textbf{LLaVA-Med} & CLIP ViT (2D) & Vicuna \cite{vicuna2023} & Linear Projection & 2D image input instead of 3D \\
    \textbf{BiomedCLIP} & CLIP ViT-B/16 & GPT-2 \cite{Radford2019LanguageMA}  & Cosine Similarity of Modality pairs &  2D image input instead of 3D\\
    \textbf{PubMedCLIP} & CLIP-ViT/B-32 & OpenAI CLIP \cite{Radford2021LearningTV} & Scaled pair-wise Cosine Similarity &  2D image input instead of 3D \\
    \textbf{MPMA} & CLIP-ViT/B & CXR-Bert \cite{10.1007/978-3-031-20059-5_1} & Global and Local Alignment & Lack of MRI VQA training\\
    \textbf{VILA-M3} & OpenAI’s CLIP-L & Vicuna \cite{vicuna2023} & Auxiliary Token Injection & High inference latency from multiple expert models \\
    \textbf{Med3DVLM} & DCFormer (3D) \cite{ates2025dcformer} & ClinicalBERT \cite{Wang2023OptimizedGC} & MLP-Mixer Projector & Lacks neuro-oncology data \\
    \textbf{MedGemma-1.5} & MedSigLIP & Gemma 3 \citep{gemmateam2025gemma3technicalreport} & Voxel Mapping & Lacks neuro-oncology data \\
    \textbf{MedImageInsight} & DaViT-360M \cite{ding2022davit} & Florence-252M \cite{yuan2021florencenewfoundationmodel} & Unified Contrastive Learning (UniCL) \cite{yang2022unified} & 2D MRI slice input instead of 3D \\
    \textbf{Lingshu} & Qwen2.5-VL \cite{bai2025qwen25vltechnicalreport} & Qwen2.5-VL \cite{bai2025qwen25vltechnicalreport} & MLP-based Projector & 2D MRI slice input instead of 3D \\
    \textbf{Merlin} & I3D ResNet152 & Clinical Longformer \cite{li2022clinicallongformerclinicalbigbirdtransformerslong}  & Cross-modal retrieval with ICD codes & Only CT support, no MRI training \\
    \textbf{CT-CLIP} & CLIP & CXR-Bert \cite{10.1007/978-3-031-20059-5_1} & Contrastive Loss &  Only CT support, no MRI training\\
    \textbf{Pillar-0} & Atlas \cite{agrawal2025atlas} & Qwen3 \cite{yang2025qwen3technicalreport}  & Asymmetric contrastive pretraining &  Lacks neuro-oncology data\\
    \textbf{CALM-VLM} & DCFormer (3D) \cite{ates2025dcformer} & Qwen-2.5-7B-Instruct \cite{qwen2025qwen25technicalreport}  & MLP-mixer architecture &  For Alzheimer's classification; not VQA\\
    \textbf{Prima} & CLIP-ViT & GPT-2 \cite{Radford2019LanguageMA} & Cosine Similarity & Not trained for VQA \\
    \hline
    
    \end{tabular}
}
\label{tab:vlms}
\end{table}

\section{Methods}

We aim to address the limitations of existing clinical VQA datasets in the context of brain MRI interpretation. Specifically, we aim to curate a dataset that combines multi-series, multi-slice 3D brain MRIs with clinically-relevant questions to enable the benchmarking of existing and new VLMs on VQA tasks for neuro-oncology. The study was conducted under an approved institutional IRB. 

\subsection{UCSF-PDGM brain MRI dataset}
University of California, San Francisco Preoperative Diffuse Glioma MRI (UCSF-PDGM) dataset \citep{doi:10.1148/ryai.220058, calabrese2022ucsfpdgm} consists of pre-operative brain MRIs for 501 patients with diffuse gliomas. The MRI studies were conducted using a standardized 3-T protocol that predominantly employed three-dimensional (3D) imaging, including diffusion and perfusion imaging for more advanced clinical interpretation. The imaging dataset is available publicly upon signing a data use agreement. 

\subsection{QA pair generation}

\begin{figure}[h]
\centering
\includegraphics[width=1\linewidth]{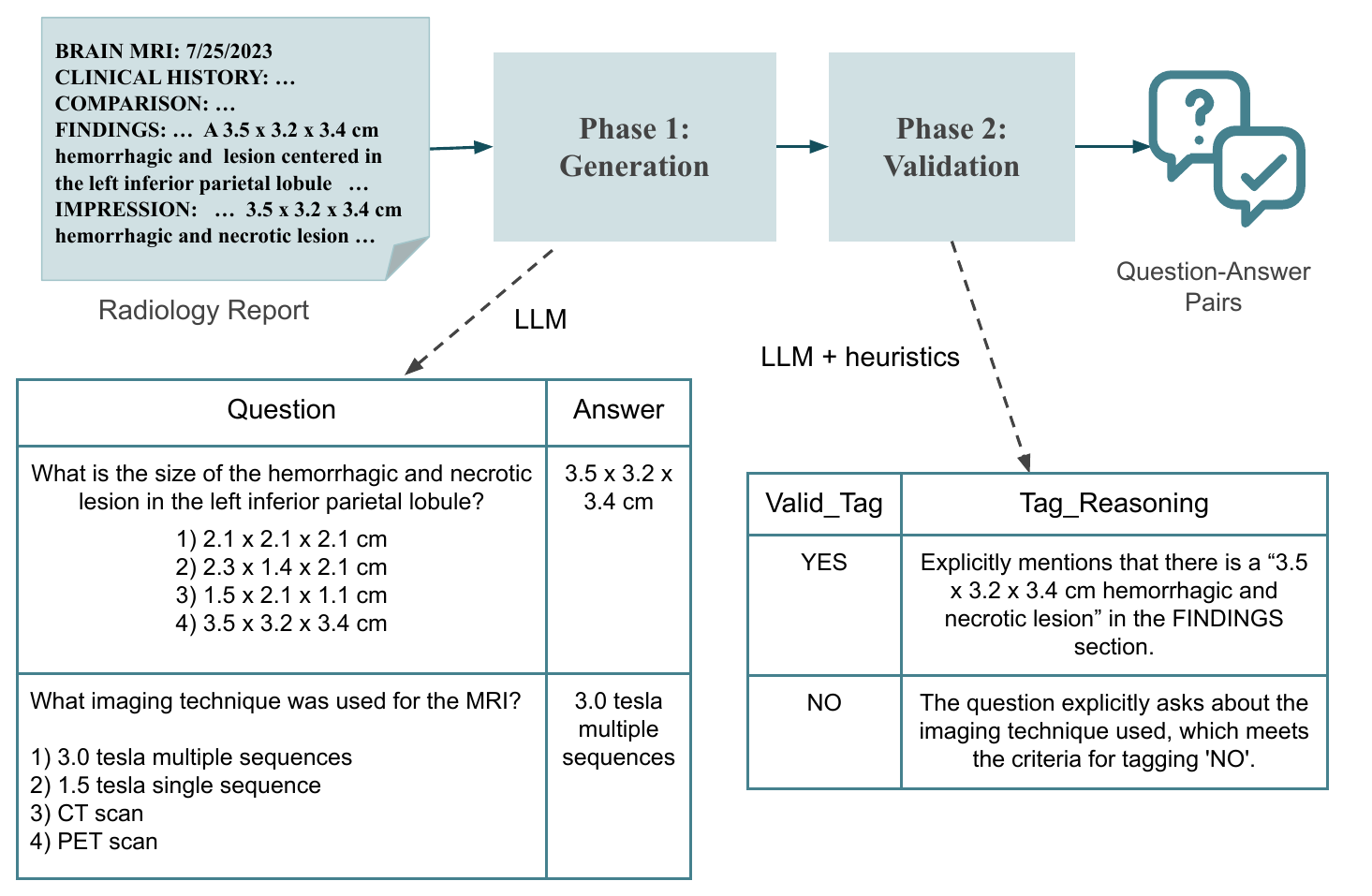}
\caption{Figure showing the data generation pipeline, along with the intermediate output in each phase. The generation phase produces candidate question-answer pairs, while the validation phase filters irrelevant or unanswerable QA pairs.}
\label{fig:qa_gen}
\end{figure}

We used the same set of brain MRI scans as those included in the UCSF-PDGM study, along with their corresponding radiology reports documenting the study's key findings and radiologic impressions, which were repurposed into multiple-choice question-answer (QA) pairs. 19 of 501 studies were excluded due to data mapping challenges. The pipeline comprised two phases: 

\begin{enumerate} 

\item Generation: A large-language model (LLM), specifically the GPT-4o model \citep{openai2024gpt4ocard}, was prompted to generate up to 20 question-answer pairs per study, with four options per question, based on the \textit{Findings} and \textit{Impression} sections of the report. The model was constrained to generate only questions whose correct answers were contained within these sections and to avoid including any unanswerable questions, such as those related to imaging technique, clinical history, disease progression, or a different anatomy, like the spine. The model was instructed to exclude options such as \textit{Not Discussed} or \textit{None of the above}, retaining only those questions that could be answered based on the imaging data. Finally, we opted for a closed-ended, multiple-choice QA setting to enable robust, automated evaluation of model capabilities. Although an open-ended QA format may be more desirable in real-world settings, it is challenging to scale clinical evaluations across multiple models, runs, and ablation settings, thus providing an incomplete view of model performance and robustness. While LLM-as-a-judge methods are popular, there are well-known limitations of these methods for model evaluation \cite{10.5555/3666122.3668142}, particularly for domains requiring advanced knowledge, such as neuro-radiology. 

\item Validation and Filtering: A separate instance of the GPT-5.2 model \citep{openaiIntroducingGPT52} was prompted to use normal reasoning effort to determine whether: (a) the generated question-answer pairs were answerable solely from the image and the report, (b) to reduce ambiguity, ensure that the laterality or location of a concerned mass or lesion within the brain was specified within the question, (c) to ensure that the questions about lesion size included the spatial directions along which the size should be reported, and (d) to rephrase any unanswerable or concerning questions. A keyword-based filter was subsequently applied to remove any question-answer pairs that contained keywords indicating unanswerable questions, such as those requiring either clinical history, temporal information, or external data to be answered correctly (e.g., postsurgical, metastasis, recurrent, spine). Duplicate questions were additionally removed through a lexical match. Manual quality validation was performed on a set of 200 QA pairs iteratively before generating the final dataset. A final subset of 75 questions was evaluated by a neuro-radiology fellow for clinical relevance and answerability.
\end{enumerate}

The full pipeline is illustrated in Figure \ref{fig:qa_gen}, the specific model settings and prompt for the generation phase are placed in Appendix \ref{appendix:generation}, the generation phase JSON output structure is in Appendix \ref{appendix:generation_structure}, the specific model settings and prompt for the validation phase are placed in Appendix \ref{appendix:validation}, and the validation phase JSON output structure is in Appendix \ref{appendix:validation_structure}. 

\subsection{Modeling Baselines}

\begin{figure}
     \centering
     \includegraphics[width=0.65\linewidth]{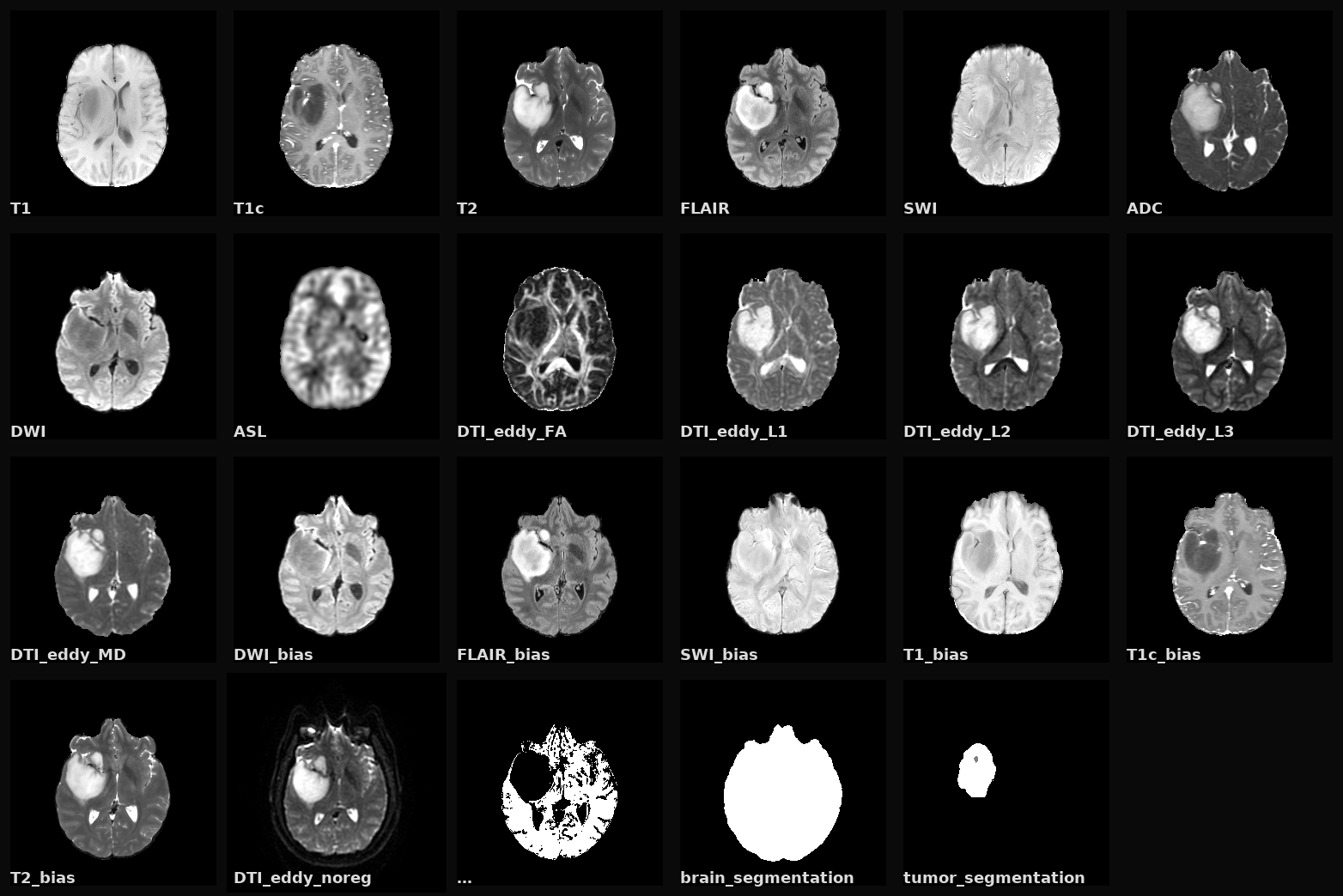}
     \caption{Example of Axial MRI montage}
     \label{fig:montage}
 \end{figure}

We evaluated the following vision-language models on the UCSF-PDGM-VQA dataset to establish performance benchmarks in a zero-shot setting: LlaVa-Med \citep{li2023llava}, MedImageInsight \citep{codella2024medimageinsightopensourceembeddingmodel}, Med3DVLM \citep{xin2025med3dvlm}, Lingshu \citep{lasateam2025lingshugeneralistfoundationmodel}, MedGemma 1.5 \citep{sellergren2025medgemma}, and the closed-weight GPT5-mini model \citep{singh2026openaigpt5card}. Since these models cannot process an entire MRI study at once, and several models also do not support multi-slice input, initial experiments evaluated different input representations for robust model performance. The most informative slices were selected as those containing the highest tumor volume, identified using a brain tumor segmentation model, the Swin-UNETR model \citep{hatamizadeh2021swin}, which was previously trained on the BRaTS dataset for brain tumor segmentation \citep{baid2021rsnaasnrmiccaibrats2021benchmark}. For models only capable of supporting a single image, axial orientation of the FLAIR scans were prioritized. Additionally, a second setting was tested, which combined the highest tumor volume slices from all axial imaging series, along with brain and tumor segmentations, into a single composite grid to provide maximal information through a single image input (Composite Montage, example in Figure \ref{fig:montage}). This enabled us to provide all key imaging slices as input models constrained to process a single scan input.

Only two models --- the MedGemma-1.5 model and GPT5-mini model --- were capable of supporting multiple image inputs, and thus they were evaluated in an additional multi-slice input configuration. Under this setting, all axial slices corresponding to the highest tumor volume, across all imaging series, were input as separate images to provide responses to the QA pair. The slices selected corresponded to the Composite Montage setting, but instead of providing them as a montage input, they were provided to the model as separate images. This created an input of up to 23 image slices per MRI study, mapped to each QA pair. We further tested a multiple montage input to these models as well, which provided as input montage composite images in the axial, saggital, and coronal orientations.

Since the UCSF-PDGM-VQA dataset is a closed-ended, multiple-choice QA dataset, model accuracy was reported as the performance metric. To establish a reliable performance bound, zero-shot inference was conducted three times, with results averaged across all runs.

\subsection{Clinical performance}
To ensure the clinical relevance of the generated dataset and to validate the LLMs's ability to produce high-quality VQA pairs, we conducted a human evaluation on a random subsample of 75 questions. These questions were provided to a neuroradiology fellow to establish an upper-bound clinical baseline to evaluate VLM architectures. Human evaluation interface consisted of four components: (1) human expert agreement on the correct answers, (2) clinical relevance of the question-answer pair, (3) the clinician's self-reported confidence score for their answer, and (4) an assessment of whether the question could be accurately answered using the provided image. This allows for quantifying the quality of the dataset, while also contextualizing VLM performance against radiology performance, providing a baseline to evaluate future model architectures. 

\subsection{Robustness tests / Ablation analysis}
To evaluate potential modality collapse and the impact of language priors, we conducted multiple ablation studies with the following settings:

\textbf{Text-only LLM baseline:} We used the Qwen3-8B LLM \citep{yang2025qwen3technicalreport} to establish a text-only baseline, comparing VLM performance against a language-only model. We opted for the Qwen model, specifically given its strong performance on diverse tasks.

\textbf{Blank image input to VLMs:}
All MRI scans were replaced with a plain black ("blank") image to evaluate the ability of VLMs to ground their responses on image input.

\textbf{Shuffled choice options:} We randomized the order of the multiple-choice options to ensure that answer selection was driven by actual understanding rather than statistical positional bias.

\section{Results}

\subsection{Dataset Statistics and Composition}
The UCSF-PDGM-VQA dataset curated in this study includes 2,387 question-answer pairs, with four multiple-choice options each, corresponding to 473 brain MRI studies. Key dataset statistics are presented in Table \ref{tab:vqa-ds}. The QA pairs span questions related to tumor size, location, anatomical changes, and tumor diagnosis. Sample QA pairs are provided in the Appendix \ref{app:samples}. During expert clinical review on a subset of 75 questions, only one question was assessed as unanswerable, and 86.7\% of the question-answer pairs were evaluated to be clinically relevant. The main dataset limitations were assessed to be the ambiguity in a very small subset of questions, such as four questions relying on inferences of the severity of the underlying conditions as \textit{mild}, \textit{moderate}, or \textit{severe}, without any standardized definitions of these terms. These challenges are reflected in human vs. model performance metrics reported in Table \ref{tab:human_eval}.

\begin{table}[h]
\centering
\caption{Key statistics for the UCSF-PDGM-VQA dataset}

\resizebox{\textwidth}{!}{%
    \begin{tabular}{|l|l|l|l|l|}
    \hline
    \textbf{Size} & \textbf{Num studies} & \textbf{Avg. words per QA} & \textbf{Avg. series per study} & \textbf{Avg. slices per series} \\
    \hline
    
    \hline
    2,387 & 473 & 188.07 & 23 & 150.69 \\
    \hline
    
    \end{tabular}
}
\label{tab:vqa-ds}
\end{table}

\subsection{Model performance}

\begin{table}[h]
\centering
\caption{Zero-shot VQA model accuracy (\%) on the UCSF-PDGM-VQA dataset. All metrics are averaged over 3 runs. The \textit{MRI} column shows the model accuracy given both the MRI and the question into the model. In the \textit{Single-Slice} setting, the MRI input is the highest tumor-volume slice from the Axial FLAIR scan. The \textit{Multi-Slice} setting includes multiple 2D slices from the MRI, representing the highest tumor volume slices from all axial sequences. In the \textit{3D} setting, the input is the full 3D Nifti volume for Axial FLAIR scan. The \textit{MRI montage} column shows the model accuracy when given both the MRI and the question, where the MRI input is a composite of the highest tumor-volume slices from all axial sequences in the MRI study, as well as containing the tumor and brain segmentation outputs. While single-slice montage refers to a composite of all axial slices, multi-slice montage refers to the use of Axial, Coronal, and Saggital montage as three inputs. The \textit{Black image} column shows the model accuracy given a blank image and the question. The \textit{Reshuffle} column shows the model's accuracy when the question choices are shuffled. For all VLMs, this setting includes the same visual input as the \textit{MRI} setting. The \textit{Text-Only} column shows the LLM accuracy when only given the question, without any visual input.}
\vspace{5px}
\resizebox{\textwidth}{!}{%
    \begin{tabular}{|l|l|l|l|l|l|l|l|}
        \hline
    \textbf{Model} & \textbf{MRI} & \textbf{MRI montage} &\textbf{Black Image} & \textbf{Reshuffle} & \textbf{Text-Only} \\
    \hline
    
    \hline
    Qwen3-8B & - & - & - & 47.73 & 52.57 \\
    LLaVA-Med-1.5 & 59.14 & 59.60 & 59.05 & 40.55 & - \\
    MedImageInsight & 20.77 & 29.69 & 18.68 & 28.01 & - \\
    Med3DVLM-Qwen-2.5-7B - 3D & 26.91 & - & 42.42 & 38.55 & - \\
    Lingshu-32B & 61.40 & 63.20 & 66.04 & 52.17 & - \\
    MedGemma-1.5-4B - Single Slice & 55.37 & 51.69 & 43.38 & 43.38 & - \\
    MedGemma-1.5-4B - Multi Slice & 63.57 & 59.08 & - & 51.16 & - \\
    GPT5-mini - Single Slice & 34.70 & 38.59 & 32.39 & 19.87 & 18.57 \\
    GPT5-mini - Multi Slice & 23.67 & 27.45 & - & 22.19 & - \\
    \hline
    
    \end{tabular}
}
\label{tab:model_eval}
\end{table}

\begin{table}[h]
\centering
\caption{Model performance on the human evaluation subset, comprising 75 QA pairs, compared to neuro-radiology fellow performance. Retaining only the most promising models based on performance on the complete dataset.}
\begin{tabular}{|l|l|l|}
\hline
\textbf{Setting} & \textbf{MRI} & \textbf{MRI montage} \\
\hline
Human & 87.88 & - \\
MedGemma-1.5-4B Single slice & 72.73 & 60.61  \\
MedGemma-1.5-4B Multi slice & 66.67 & 66.67  \\
Lingshu-32B & 57.07 & 64.14\\
Qwen3-8B LLM & 53.03 & - \\

\hline
\end{tabular}
\label{tab:human_eval}
\end{table}

Table \ref{tab:model_eval} provides the average zero-shot accuracy of the evaluated models on the UCSF-PDGM-VQA dataset. Similarly, Table \ref{tab:human_eval} contextualizes model performance against clinical performance on a subset of 75 QA pairs. Evaluations on the UCSF-PDGM-VQA dataset reveal a significant performance gap; among the models tested, the MedGemma-1.5 model, within multiple-slice and single-slice settings, achieved the highest accuracy at 63.57\% and 55.37\%, respectively. Performance improvement of nearly 8\% through multiple slice inputs suggests that the model relies on additional sequences for accurate inference, which also aligns clinically with the expected behavior. MRI montage input harmed the model performance, indicating its inability to process composite images. Even within the highest-performing models, the gap compared to the clinical performance is large (nearly 15\%), with the neuro-radiology fellow performance upper bound on the human evaluation subset being nearly 88\%. This highlights the limitations of specialist medical VLMs in interpreting multi-sequence, multi-slice brain MRI inputs. Although the Linghsu and LlaVa-med models seem to be performing similarity, ablation settings are concerning and discussed later. The GPT5-mini model struggled to make inferences from multiple images as the input, demonstrating large performance drops compared to single slice settings. Even within single slice or montage settings, the model performance was quite low, which demonstrates the lack of reliable performance for brain MRI inference. The MedImageInsight and the Med3DVLM models, despite being customized for medical data, only performed at random for brain MRI VQA, additionally demonstrating the lack of their inference capability in this domain.

Within the ablation settings, we noticed very interesting findings. The Lingshu model and the LLaVA-Med model, despite their higher performance in the MRI and MRI montage settings, performed similarly when given a black image as input. These results highlight a strong bias toward language priors and the ordering of options in multiple-choice QA settings. Of note is that the Lingshu model includes brain MRI data in its model training data, and the lack of generalizability stems despite that. Furthermore, although the Med3DVLM model performed at random with MRI input, its performance improved significantly when the MRI input was removed or when the option order was shuffled, thus indicating its inability to process brain MRI scans. This is expected since the model was not trained on brain MRI data. MedGemma model performance dropped when provided with black image inputs, indicating its reliance on input MRI data. The GPT5-mini model, however, performed similarly when provided with a black image as input as compared to a single FLAIR MRI slice, highlighting its inability to process that single slice effectively.

\begin{table}[htbp]
    \centering
    \caption{Example of Questions LLM got correct when provided no image}
    \label{tab:location_questions_qwen}
    \begin{tabular}{|p{0.9\textwidth}|}
        \hline
        \multicolumn{1}{|c|}{\textbf{Questions}} \\
        \hline
        What are the locations involved by the multifocal hemorrhagic high-grade glial neoplasm? \\
        \hline
        What are the locations of the two synchronous masses (presumed diffuse glioma)? \\
        \hline
        What is the anterior location of the tumor’s extension within the temporal lobe? \\
        \hline
        What is the draining location of the large developmental venous anomaly (DVA) at the superomedial aspect of the right temporal lobe tumor? \\
        \hline
        What is the imaging extent/location of the surrounding edema? \\
        \hline
        What is the location and extent of the abnormal infiltrative T2 signal? \\
        \hline
        What is the location of mucosal sinus thickening? \\
        \hline
        What is the location of the abnormal FLAIR signal? \\
        \hline
    \end{tabular}
    \centering
\end{table}

Moreover, Qwen3-8B  text-only model, without any image input, surprisingly demonstrated strong performance when provided only with the question and options. At nearly 50\% accuracy, the models performed twice as well as random, matching the performance of the best VLM under reshuffled input settings. Table \ref{tab:location_questions_qwen} provides a subset of the questions that Qwen3-8B was able to answer correctly despite its lack of image input. This finding indicates a strong prior in LLMs towards question-answering tasks, as additionally reflected in other VLM ablation experiments, such as the use of blank image input. This suggests that LLMs contribute more strongly towards VLM outputs compared to vision inputs themselves. Interestingly, the text-only inference within the GPT5-mini model did not demonstrate the same pattern, performing poorly even compared to random baselines. However, the model performance even in VLM settings is low, thus suggesting its inability to perform inferences related to brain tumor MRI more generally.

Finally, across most models, we observed large (often 10\% or more) performance drops when shuffling the order of options in the question. This suggests a strong bias in the models towards the position of the correct answer. This may be expected given that the questions and their options were generated using an LLM. We further investigated a potential order bias in the options in the generated dataset. We identified that the first option corresponded with the correct answer 84.7\% of the time, option 2 --- 12.2\%, option 3 ----- 2.7\%, and option 4 was correct only in 0.5\% of the questions. This highlighted two major vulnerabilities: (a) LLMs, specifically the GPT-4o model, generated the correct option first before adding incorrect options, and (b) vision-language models were able to exploit this vulnerability even in zero-shot inference settings. In the final public version of the dataset, we provide both shuffled and unshuffled options, with the shuffled case as the default for inference, to enable robust benchmarking while also enabling reproducibility.

\section{Discussion}

Popular clinical vision-language models have leveraged and adapted general vision encoders to better capture key regions in medical images, while adjusting their vision-text fusion strategies to improve the model's understanding of the input data. However, even with these architectural changes, challenges remain with the VLM's comprehension of domain-specific knowledge and with correctly encoding medical images and fusing them with text tokens. Specifically, current models are limited in their ability to process 3D volumes natively, with most expecting only a limited number of 2D slices as input. This poses a significant constraint for domains such as neuro-oncology, where accurate interpretation relies on a joint inference of multiple 3D imaging sequences at once, which the current models are incapable of processing. This is, in turn, reflected within model performance on real-world benchmarks for brain MRI interpretation. In this study, we developed a new VQA dataset specific to neuro-oncology, pairing multi-sequence brain MRI scans with clinically relevant question-answer pairs to enable benchmarking of VLM performance in realistic clinical settings. Through this dataset, we identified significant gaps in current model capability. Even in closed-ended QA settings, the best models achieved only 64\% accuracy, demonstrating a significant gap compared to both clinical needs and the clinical performance upper-bound. This performance is expected to only worsen in open-ended settings. Poor accuracy is driven in part by the lack of models capable of encoding multi-sequence 3D MRI volumes for reliable brain interpretation, and in part by their inability to leverage visual evidence and over-reliance on linguistic patterns.

As also discussed previously by Asadi et al. \cite{asadi2026mirage}, current VLMs are susceptible to modality collapse, especially in medical settings. While the previous study identified these limitations in the context of existing clinical vision-language datasets for tasks other than VQA, our study confirmed modality collapse during VQA for interpreting glioma MRIs. Although during clinical evaluations, it was noted that the imaging data was necessary for model inference, most models performed similarly with and without imaging input. Both improved and at-par performance of several models when using a blank image input or a text-only LLM suggests that the VLMs are not reasoning well enough with the MRIs to draw meaningful analysis from the input images, or do not consider the images useful enough when answering the question. Some examples where we witnessed modality collapse included questions related to tumor size, tumor location, and the underlying diagnosis, which are all critically dependent on an individual's MRI scans. The fact that models can answer these questions without vision inputs is deeply concerning. This underscores the need for model architectures that enforce strict visual grounding and reduced reliance on language priors for grounded, accurate inference over brain MRI scans. Inability to overcome modality collapse, and yet obtaining high performance on tasks that are clinically unanswerable without underlying imaging data, will result in dangerous clinical hallucinations and put patients at a critical safety risk. 

\section{Conclusions}
We created a clinically relevant benchmark of 2,387 closed-ended visual question-answer pairs, corresponding to 497 multi-series 3D brain MRI studies, and released it publicly to enable future studies. Through an analysis on this benchmark, we identified: (a) significant performance gaps between existing VLMs and the minimum performance required to enable practical use, thus highlighting a need to improve models for brain MRI VQA, and (b) modality collapse, a key limitation of currently popular medical vision-language models in closed-ended VQA over brain MRIs. Insufficient encoding of volumetric MRI data prevents existing models from effectively leveraging critical information in MRI scans and mapping them to the corresponding text inputs, leading to incorrect VQA inference. This raises a key safety challenge for the potential future deployment of VLMs in clinical settings.

\section{Limitations and Future Directions}
Key limitations of this study are the reliance on imaging data from a single institution, for a single radiologic modality (MRI), for a single disease group (diffuse gliomas), a single anatomy (brain), and a single time point (pre-operative). Thus the results can only be interpreted within these contexts. Furthermore, although we opted for a multiple-choice setting in this study to enable automated evaluation in clinically relevant task setups, the eventual goal is to enable a pathway towards multi-turn conversations within radiologic settings, where users can interact with a vision-language model in an open-ended manner to assist with speedier MRI analysis. To this end, future research will incorporate multi-turn, open-conversational settings, with access to longitudinal scans and patients' clinical history, thus simulating realistic dialogue alongside reasoning-driven responses, as opposed to multiple-choice settings proposed in the current study.

\bibliographystyle{plainnat}
\bibliography{references}

\newpage

\appendix
\section{QA Pair Generation Settings and Prompt}
\label{appendix:generation}

For the generation phase, we adjusted the GPT-4o LLM to have a temperature of 0.15 to increase the strictness of this process and prevent the LLM from deviating from the guidelines in the prompt. We accessed this model via the Versa API to ensure data security. 

The generation prompt is as follows: 
\begin{lstlisting}[language=prompt] 
You will be given the radiology report of a patient. Your job is to create question-answer pairs based on the information given in the `IMPRESSION' and `FINDING' sections of the report. 

Each question MUST have 4 options, 1 correct option and 3 incorrect options, and these options MUST be in the same string as the question. You MUST create 20 question-answer pairs

EXAMPLE QUESTION FORMAT:
Where is the location of the tumor?

1) Upper Left Region  2) Upper Right Region  3) Lower Left Region  4) Lower Right Region
    
EXAMPLE ANSWER FORMAT:
Upper Right Region

    
CREATE QUESTION-ANSWER PAIRS BASED ON THE INFORMATION BELOW:
- Please answer the following list of questions and provide the reasoning for each answer. 
- Please format the response so that the reasoning is clearly separated from the answer. 
- Place the reasoning section before the answer.  
- Please quote direct full sentences of evidence from the report in the reasoning section to help justify the answer. 
- Each question will provide the multiple options of the answer, pick one of them and follow the instructions on how to answer. 
- An answer of "no" means that the report specifically confirms the answer to the question is no and there is clear evidence to confirm this. 
- A reasoning of "inconclusive" means "insufficient conclusive evidence" or that there might be some evidence to indicate some answer, but there isn't enough to confidently conclude an answer. 
- An answer of "not discussed" means "not discussed in the report" or that the question topic was not mentioned in the report at all. Keep the original numbering for the list of questions.
- DO NOT include any questions that are not related to the "IMPRESSION" and "FINDINGS" sections.
- DO NOT include any follow-up questions or questions that REQUIRE knowledge outside of the report
- DO NOT include any questions that ask about `residual' portions of the tumor or questions about a previous MRI
- DO NOT use the phrase `in the report' in the questions. The questions should be answerable with only the MRI.
- ALL ANSWERS should be in the text and should never be "None of the Above"
    
RADIOLOGY REPORT:
(radiology report)
\end{lstlisting}

\newpage
\section{QA Pair Generation Output Structure}
\label{appendix:generation_structure}

The generation phase JSON output structure is as follows:
\begin{lstlisting}[language=prompt]
{
    `question' : str,
    `answer' : str,
    `reasoning' : str
}
\end{lstlisting}

\newpage

\section{QA Pair Validation Prompt}
\label{appendix:validation}

For the validation phase, we adjusted the GPT-5.2 LLM to have a temperature of 0.15 to increase the strictness of this process and prevent the LLM from deviating from the guidelines in the prompt. We accessed this model through the Versa API for data security.

The validation prompt is as follows:
\begin{lstlisting}[language=prompt] 
Given a radiology report for a brain MRI, please go through each of the question-answer pairs and determine if the pairs can be answered given the criteria below. 

Please answer the following list of questions and provide the reasoning for each answer. 
Please format the response so that the reasoning is clearly separated from the answer. 
Place the reasoning section before the answer. 
Please quote direct full sentences of evidence from the report in the reasoning section to help justify the answer. 
Each question will provide the multiple options of the answer, pick one of them and follow the instructions on how to answer.
Keep the original numbering for the list of questions.
If the question-answer pairs MEETS any of the criteria below, then tag them with the "NO" string.
If the question-answer pairs does NOT MEET any of the criteria below, then tag them with the "YES" string.
Also, be sure to explain why you chose the tag in the `tag reasoning' response.
    
CRITERIA:
- ANY question-answer pairs that require the patient's clinical history, previous brain MRIs, or any other information outside of the report to answer (Questions about "midline shift" are ok and should NOT be tagged `NO')
- ANY question-answer pairs with the reasoning of "Inconclusive" and nothing else
- ANY question-answer pairs with the answer of "Not discussed" and nothing else
- ANY questions that explicitly asks to compare the MRI with a previous MRI or ask about a previous MRI. (Some keywords: postsurgical changes, postsurgical, retrospect, progression, recurrent, stable, tumor growth, tumor shrinkage, metastasis)
- ANY questions that REQUIRE knowledge outside of the report to answer it.
- ANY questions that ask about `residual' portions of the tumor.
- ANY question-answer pairs with the answer of "None of the above".
- ANY question-answer pairs where it asks what technique is being used in the report in an explicit or implicit manner. (Some keywords: multivoxel spectroscopy, FLAIR, T1, T2)
- ANY questions that are not related to the brain  
- ANY questions that are not about aspects found in the brain MRI

You also have the ability to change questions if they do not meet the QUESTION CHANGING CRITERIA below.
Your job is to look over the input question-answer pair and make the changes to the questions and answers based on the information given in the `IMPRESSION' and `FINDING' sections of the report.
ONLY MAKE CHANGES if the question-answer pair meets the criteria below that show which questions need to be changed and how they should be changed, otherwise keep everything the same.
Do NOT add `based on the report' in any of the questions.
    
Your outputted question MUST contain a new question or the original question.
Your output choices MUST include FOUR potential choices. ONE should be the answer based on the report, the others should be changed to more easily differentiate the incorrect choices from the correct one, or be the same as the original choices.
Your outputted answer MUST be ONE of the potential choices and must be based on the radiology report.

QUESTION CHANGING CRITERIA:
- Any questions asking about the size MUST specify what dimensions it is looking for. You MUST add the dimension format used to answer the question 
(EX: DIMENSION: (x, y, z) for 5 x 5 x 5 cm. DIMENSION: (x, y) for 5 x 5 cm) 
Be sure to space out the potential choices so only one choice is correct within a margin of error (For cm measurements, you MUST have a 1 cm difference between choices. For midline shifts and bigger structures, you MUST have a 5mm difference between choices. For smaller structures, like pituitary gland, you MUST have a 3mm difference between choices.) 
- Any questions that are asking about a specific aspect (e.g, lesion, mass, tumor, anything dependent on anatomy) of the MRI MUST be sure to change the question so we know the exact location of where the characteristic is. You MUST be as descriptive as possible when describing the location. 
If the location of the specific aspect is unknown, then you MUST include the aspect's laterality.
  
QUESTION-ANSWER PAIRS:
(QA pairs from the generation phase)
\end{lstlisting}

\newpage
\section{QA Pair Validation Output Structure}
\label{appendix:validation_structure}

The validation phase JSON output structure is as follows:
\begin{lstlisting}[language=prompt]
{
    `question' : str,
    `answer' : str,
    `reasoning' : str,
    `tag' : str,
    `tag_reasoning' : str
}
\end{lstlisting}

\section{Sample QA Pairs}
\label{app:samples}

\begin{table}[h]
\centering
\caption{Sample question-answer pairs in the curated UCSF-PDGM-VQA dataset.}
\vspace{5px}

\resizebox{\textwidth}{!}{%

    \begin{tabular}{|p{0.8\linewidth}|l|}
        \hline
    \textbf{Question} & \textbf{Answer} \\
    \hline
    
    \hline
    What is the location of the cystic lesion?
1) Left frontal lobe
2) Right retrolenticular internal capsule/thalamus
3) Left parietal lobe
4) Right occipital lobe & Right retrolenticular internal capsule/thalamus \\ \hline
    What is the observed mass effect on the right lateral ventricle? 
1) Compression
2) Expansion
3) No effect
4) Displacement & Compression \\ \hline
    What is the measured midline shift (right-to-left) due to mass effect?
1) 9 mm right to left
2) 4 mm right to left
3) 14 mm right to left
4) 7 mm left to right & 9 mm right to left \\ \hline
    What is the size of the lesion in the left frontal middle gyrus? DIMENSION: (AP x transverse x CC)
1) 3.1 x 2.9 x 3.8 cm
2) 2.1 x 1.9 x 2.8 cm
3) 4.1 x 3.9 x 4.8 cm
4) 1.1 x 0.9 x 1.8 cm & 3.1 x 2.9 x 3.8 cm \\ \hline
    What is the observed vascularity of the mass centered in the posterior right temporal lobe?
1) Significant hypervascularity
2) Minimal vascularity
3) No vascularity
4) Moderate vascularity
 & Significant hypervascularity \\ \hline
    Which structure is specifically noted to have surrounding T2 FLAIR hyperintensity extension?
1) Right optic tract
2) Left optic tract
3) Right cerebellar hemisphere
4) Left cerebellar hemisphere & Right optic tract \\ \hline
    What is the size of the heterogeneously enhancing lesion centered in the right insula and temporal stem? DIMENSION: (AP, transverse)
1) 4.5 x 3.1 cm
2) 5.5 x 4.1 cm
3) 3.5 x 2.1 cm
4) 6.5 x 5.1 cm
 & 4.5 x 3.1 cm \\ \hline
    What is the degree of midline shift observed?
1) 2 mm
2) 6 mm
3) 11 mm
4) 16 mm
 & 6 mm \\ \hline
    What is the size of the mass centered within the left frontotemporal lobes (including the left hippocampus) with involvement of the left insula? DIMENSION: (x, y) 
1) 4 x 4 cm
2) 5 x 5 cm
3) 6 x 6 cm
4) 7 x 7 cm & 6 x 6 cm  \\ \hline
    What type of enhancement does the large mass centered in the left frontotemporal lobes with involvement of the left insula demonstrate?
1) Thin rim enhancement
2) Thick rim enhancement
3) No enhancement
4) Uniform enhancement
 & Thick rim enhancement \\ \hline
    What is the observed left-to-right midline shift associated with the left frontotemporal-insular mass?
1) 0.5 cm
2) 1.0 cm
3) 1.5 cm
4) 2.0 cm
 & 1.5 cm \\
    \hline
    
    \end{tabular}
}
\end{table}

\section{Model Evaluation hyperperameters and set up}\label{sec:hardware_selection}

\begin{table}[t]
\centering
\caption{Model configurations and inference settings for medical VQA evaluation.}
\resizebox{\textwidth}{!}{%
\begin{tabular}{llll}
\hline
\textbf{Model} & \textbf{Key Inference Settings} \\
\hline
MedGemma-1.5-4B  & Greedy decoding (do\_sample=False), max\_new\_tokens: 512, bf16/fp16, Structured Output (Pydantic) \\
GPT5-mini  & max\_completion\_tokens: 4096, Structured Output (Pydantic), "gpt-5-mini-2025-08-07" \\
Lingshu-32B  & bf16, max\_new\_tokens: 512, SDPA attention implementation, Structured Output (Pydantic), \\
LLaVA-Med-1.5  & temperature=.2, do\_sample=True \\
MedImageInsight  & CLIP-style zero-shot similarity (no text generation) \\
Med3DVLM  & Greedy decoding (do\_sample=False), max\_new\_tokens: 512, temperature=0 \\
Qwen3-8B  & Greedy decoding (do\_sample=False), \\
\hline
\end{tabular}
}
\label{tab:hyperparam}
\end{table}

All experiments were conducted on an internal high-performance computing cluster equipped with 6 NVIDIA H100 and L40S  GPUs and a slurm-based scheduler. The zero-shot inference pipeline for the 2,387 QA pairs across all model configurations (Single-slice, Multi-slice, and Montage) took approximately 818 total GPU hours. Pre-processing of the MRI volumes, including brain tumor segmentation using the Swin-UNETR model, required an additional 48 hours on the same internal cluster. All model hyperparameters are reported in Table \ref{tab:hyperparam}.

\newpage

\end{document}